\documentclass[runningheads]{llncs}
\usepackage{graphicx}
\usepackage{color}
\begin{document}
\title{Local Style Awareness of Font Images}

\author{Daichi Haraguchi\and
Seiichi Uchida}

\authorrunning{D. Haraguchi et al.}

\institute{Kyushu University, Fukuoka, Japan}

% \author{Daichi Haraguchi\inst{1,2}\and
% Seiichi Uchida\inst{1,3}}
% %
% \authorrunning{Haraguchi et al.}
% \institute{Kyushu University, Japan \and
% \email{daichi.haraguchi@human.ait.kyushu-u.ac.jp} \and 
% \email{uchida@ait.kyushu-u.ac.jp}\\
% }

%
\maketitle         
\begin{abstract}
When we compare fonts, we often pay attention to styles of local parts, such as serifs and curvatures. This paper proposes an attention mechanism to find important local parts. The local parts with larger attention are then considered important. The proposed mechanism can be trained in a quasi-self-supervised manner that requires no manual annotation other than knowing that a set of character images are from the same font, such as {\tt Helvetica}. After confirming that the trained attention mechanism can find style-relevant local parts, we utilize the resulting attention for local style-aware font generation. Specifically, we design a new reconstruction loss function to put more weight on the local parts with larger attention for generating character images with more accurate style realization. This loss function has the merit of applicability to various font generation models. Our experimental results show that the proposed loss function improves the quality of generated character images by several few-shot font generation models.   

\end{abstract}
%
%
%
% ===================================================================
\section{Introduction\label{sec:introduction}}
% ===================================================================
\begin{figure}[t]
    \centering
    \includegraphics[width=0.8\linewidth]{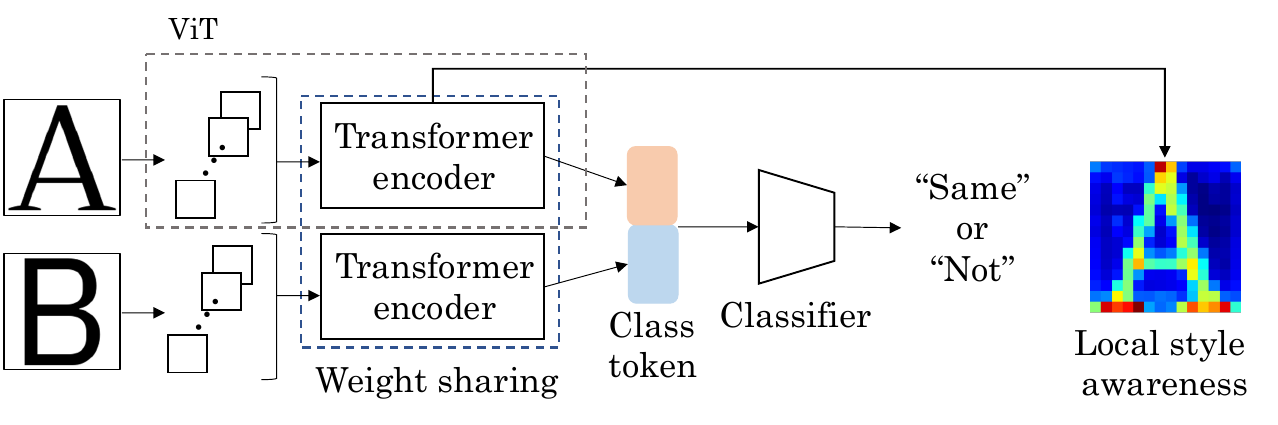}\\[-3mm]
    % \caption{ Note that input images are decomposed to patches before inputting the Transformer encoder.}
%    \label{fig:identification}
%    \includegraphics[width=0.9\linewidth]{figs/teaser.pdf}
    \caption{Overview of the proposed technique to determine {\em local style awareness}, which indicates important local shapes to describe font styles. The local style awareness is obtained as a fine attention map through a contrastive learning scheme that identifies whether two given character images belong to the same font.
    %The style is particular to a font and is defined by its difference from the other font. %The proposed technique, therefore, consists of a contrastive learning scheme to identify whether two given character images belong to the same font. This identification task trains the neural network model (ViT) to catch important local parts, i.e., local style awareness, as a fine attention map. Note that this contrastive learning can be performed in a quasi-self-supervised learning scheme; it does not need manual annotation except for preparing a set of multiple fonts.
    }
    \label{fig:teaser}
\end{figure}

% topic: local style awarenessについて，
To understand font styles, a reasonable choice is to observe local shapes. Each character has a shape representing font style; however, the whole character shape is unnecessary to understand its style. Assume that a character `A' is printed with {\tt Helvetica} (a famous sans-serif font), and we want to understand the style of {\tt Helvetica} from it. In this case, we must ignore the global shape that makes `A'  as `A.' In other words, we need to focus on local shapes, such as serifs, corners, stroke width, and local curvatures, which are rather independent of character class `A.'\par
Through a {\em contrastive learning} scheme, this paper tries to determine {\em local style awareness} representing important local shapes for particular font styles. Imagine a person who has only seen {\tt Helvetica} in its lifetime --- then, the person cannot determine the style of {\tt Helvetica}. In other words, we can understand the particular style of {\tt Helvetica} by contrasting (i.e., comparing) it with other fonts, such as {\tt Times New Roman} and {\tt Optima}. Moreover, as noted above, we need to ignore the whole letter shape and focus on local shapes during the comparison in some automatic way.\par
Fig.~\ref{fig:teaser} shows the overall structure of the proposed model to determine local style awareness. This figure also shows a heatmap representing local style awareness for the input `A.' This model is trained via a {\em font identification} task. This task aims to determine whether two given character images come from the same font\footnote{We can find another type of the font identification task, where a single image is given, and a model chooses its font name, such as {\tt Helvetica}, from the prespecified font classes. In contrast, our font identification task is more general, and its model answers ``Same'' or ``Not'' for a given pair of images.}. In the case of this figure, a serif-style `A' and a sans-serif `B' are given, and therefore, the model must answer ``Not.''  To answer this task correctly, the model needs to ignore the whole shapes of `A' and `B' and enhance their local style differences; therefore, the model needs to determine the local style awareness internally. By visualizing this internal representation as a spatial map, we will have local style awareness.\par 
The following two points must be considered for determining local style awareness via the font identification task. First, the task is formulated as a contrastive learning task. As noted above, font style is determined by comparing the target font with other fonts. Therefore, the model of Fig.~\ref{fig:teaser} is trained to enhance local style differences. Second, we need to compare two different alphabets, such as `A' and `B,' instead of the same alphabet, such as `A' and `A.' If we only compare the same alphabet in the font identification task, it reduces to a trivial task. The model can give perfect identification results by checking whether two inputs are entirely the same or not. In other words, the model cannot learn the local style awareness. By training the model with font pairs including different alphabets, the model can learn local style differences while ignoring the global letter shapes.\par
The proposed technique uses Vision Transformer (ViT)~\cite{dosovitskiy2020image} by expecting the merits from its attention mechanism. In ViT, a character image is decomposed into small patches, and these patches are fed into a transformer encoder. In the encoder, the attention of each patch is calculated by using the mutual relationship between patches. By contrastive learning for the font identification task, ViT will give larger attention to the local patches that are more important for representing the style. The heatmap of the local style awareness of Fig.~\ref{fig:teaser} is an attention map given by the proposed technique, and each element of the heatmap corresponds to a patch.\par
% アノテーションフリーな話
It should be emphasized that the ViT-based model of Fig.~\ref{fig:teaser} is trainable very efficiently in {\em a quasi-self-supervised manner} for the font identification task. The ground truth for our task is whether two character images are from the same font or not. Accordingly, if we prepare the character images from specific font sets, we know the font name of each image and give the ground truth for each image pair without any manual annotation cost. For example, if we prepare font sets of {\tt Helvetica} and {\tt Optima}, the pair `A' and `B' from {\tt Helvetica} should have the ground truth of ``Same,'' and the pair `C' from {\tt Helvetica} and `D' from {\tt Optima} have ``Not.'' Our experimental results show that the attention learned in this efficient manner becomes larger around important local shapes for individual styles, as expected.\par
\begin{figure}[t]
    \centering
    \includegraphics[width=0.8\linewidth]{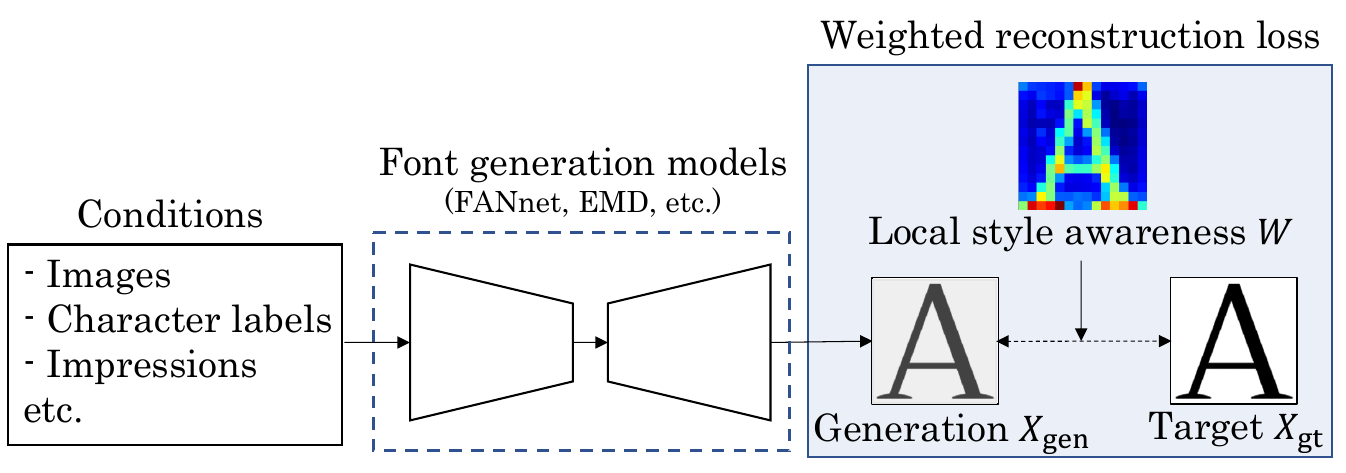}\\[-3mm]
    \caption{Training the font generation model using the reconstruction loss weighted by local style awareness.}
    \label{fig:gen_example}
\end{figure}

% topic: application,  local awareを初めて使う
In this paper, we further utilize this attention mechanism to realize local style-aware font generation models.
Specifically, as shown in Fig.~\ref{fig:gen_example}, we utilize the local style awareness representing the importance of individual patches for weighting the reconstruction loss function in various font generation models.  
This weighting scheme contributes to a more accurate reproduction of the important local parts in the generated images, as proved in our experiments of few-shot font generation in three different font generation models. 

Our contributions are as follows.
\begin{itemize}
    \item To determine local style awareness, we propose an efficient contrastive learning framework to solve the font identification task. Through the solution of the task, our network model can determine the local parts important to describe the font style. Note that the model can be trained without any manual annotation. 
    \item We experimentally proved that the above framework could determine the important local parts.
    \item We apply the local style awareness to the weight of reconstruction loss in the font generation model. This weighting scheme can be easily introduced in any model trained with a reconstruction loss. 
    \item Our experimental results show that the weighting scheme improves the quality of few-shot font generation. 
\end{itemize}

% ===================================================================
\section{Related Work}
% ===================================================================
% -------------------------------------------------------------------
\subsection{Font recognition and identification}
% フォント分類は昔からやられてきた
% Font recognition has been conducted since the 1990s.
Font recognition is a famous font-related task that is also conducted in the DAR community.
There are some pioneering works in font recognition~\cite{shi1997font,Zramdini1998}.
Shi \textit{et al.}~\cite{shi1997font} recognized the font families on documents utilizing page properties and short words in the document images.
Zramidin \textit{et al.}~\cite{Zramdini1998} used 280 fonts and recognized them using a Bayesian classifier.
Following these researches, many font recognition has been studied~\cite{aviles2005high,sun2006multi,solli2007fyfont}.\par
% VFRの話
In contrast to the above studies, which address only a small number of fonts and are weak to noise, 
Chen \textit{et al.}~\cite{chen2014large} addressed Visual Font Recognition (VFR), which is the task of recognizing font on an image or photo, and they recognized more than 2,000 fonts.
Following this, DeepFont,~\cite{wang2015deepfont} based on deep learning, was proposed for solving VFR.
Wang \textit{et al.}~\cite{wang2018font} conducted VFR for Chinese fonts using deep learning.\par
% 原口がfont identificationを提案
Haraguchi \textit{et al.}~\cite{haraguchi2020character} tackled font identification, identifying whether a pair of fonts are identical. Font recognition deals with fonts registered in advance; in contrast, font identification deals with pairs of arbitrary fonts, even test-time. 
We utilize font identification to obtain local style awareness of not only training images but also testing images.\par

\subsection{Font generation}
% \subsubsection{Evaluation for font generation}
% フォント生成が盛んに行われている
% 特に，few-shot フォント生成
% しかし，細部デザインは忠実では無い
% フォントは細部の違いで印象が変わるため問題である．
In recent years, many researchers tackled font generation, especially few-shot font generation~\cite{kong2022look,tang2022few,liu2022xmp,gao2019artistic,srivatsan2019deep,zhang2018separating}.
Few-shot font generation is a task that accepts content images or labels (i.g., a character label) and source images for extracting the style and then generating font images with the content and the style.
Most of the few-shot font generation approaches inadequately handle aesthetic details in fonts.
Fonts have their impression in their details (local structures)~\cite{ueda2021parts}; therefore, the aesthetic details in fonts are essential.\par
% 中国語では部首を活用することで,local awareな生成を試みている
% しかし，その部首に潜む細かいフォントの特徴まではみていない
% そもそも，英語ではlocal awareな生成方法なし
Some font generation for Chinese characters studies seeks the local-aware font generation to utilize radical information or structure of characters~\cite{kong2022look,tang2022few,liu2022xmp}.
However, the more detailed font styles in the radicals are not addressed.
Additionally, there is no study of local-aware font generation for alphabets.\par
% 文字をシャープにする方法や文字領域と背景領域のインバランスに対応するものはいくつかあり
Some studies address the imbalance between character regions (foreground) and the background or sharpness of characters~\cite{zhang2018separating,srivatsan2019deep}.
In our experiment, we use these approaches for comparative methods; therefore, we describe the details of these approaches in Section~\ref{sec:baseline}.\par

\subsection{Fine-grained tasks}
Fine-grained image recognition and classification focus on learning subtle yet discriminative features.
Some studies utilize attention maps to extract such features~\cite{zheng2017learning,fu2017look,zheng2019looking,zhuang2020learning}.
These methods estimate attention maps that localize the discriminative regions through end-to-end training for fine-grained image recognition or classification.
And then, they utilize the attention map to emphasize the discriminative features.
They do not need extra annotations for the regions; however, they need additional branches to estimate the attention map in each model.
Therefore, they need to propose a model for each task to obtain and utilize an attention map.\par
We obtain the attention map of local style awareness through font identification.
In contrast to the above fine-grained recognition and classification, we utilize the attention map for font generation tasks independent of font identification.
Therefore, we do not need to prepare the different models to obtain the attention map for each font generation method.\par
%
% In other local-aware tasks, especially a segmentation task, some studies proposed local-aware loss, especially for boundary-aware loss, which is valid for any backbone neural network model to improve segmentation quality~\cite{wang2022active,borse2021inverseform}.
% Borse \textit{et al.}~\cite{borse2021inverseform} proposed boundary-aware loss, which minimizes transformation parameters between the target and the estimated boundary. The transformation parameters are estimated by a neural network model independent of the segmentation model; therefore, the loss is easily introduced to any segmentation model.
% Wang \textit{et al.}~\cite{wang2022active} proposed an active boundary loss, progressively adjusting the estimated boundary to the target boundary. This loss is also easily introduced to any segmentation model.\par
% %
% We use the attention maps to the loss for the local style-aware font generation.
% Similar to the above segmentation tasks, our loss can be easily introduced in any model with reconstruction loss.

% ===================================================================
\section{Local Style Awareness in Font Images}
% ===================================================================
% -------------------------------------------------------------------
\subsection{Methodology~\label{sec:tech-details}}
\subsubsection{Font identification by contrastive learning}
To determine important local parts for font styles, 
we realize a font identification model by contrastive learning. 
As noted in Section~\ref{sec:introduction}, font styles are defined by comparing various fonts and enhancing their differences. Font identification is the task of determining such differences between two input images by comparing them in a contrastive manner. Therefore, solving the font identification task fits our aim to determine local style awareness.\par
Fig.~\ref{fig:teaser} shows the ViT-based model for the font identification task. A pair of character images are prepared, and each is fed into a ViT, i.e., a transformer encoder, after decomposing into small patches. Each ViT outputs a feature vector called a class token. A pair of class tokens are concatenated and fed to a classifier consisting of fully connected layers to make the binary decision, ``Same'' or ``Not.''\par
% -------------------------------------------------------------------
\subsubsection{Determining local style awareness by attention}
ViT, or transformer encoder, has a patch-wise self-attention mechanism, which evaluates the mutual relationship not only between neighboring patches but also between {\em distant} patches. This mechanism is useful for acquiring local style features because the style-aware local parts, such as serifs, often exist at distant locations. For example, serifs of `I' exist at the top and bottom of the vertical stroke. By training the model of Fig.~\ref{fig:teaser} for style identification, this self-attention mechanism is expected to be more sensitive to the local style difference and less sensitive to the global shapes that make, for example, `A' as `A.'\par
Accordingly, if we measure the value of patch-wise self-attention, we can get local style awareness as an attention map. (If an image is decomposed into $M\times N$ patches, the map has $M\times N$ resolution.)  Roughly speaking, in the task of font identification, the attention value will become higher (or lower) at patches that are important (or unimportant) for the identification. In our model of Fig.~\ref{fig:teaser}, we have two $M\times N$ self-attention maps corresponding to two image inputs. The attention map for each image will show the local style awareness of the image.\par
To measure the attention values, we use {\em attention rollout}~\cite{abnar-zuidema-2020-quantifying}. Attention rollout is an XAI technique and can visualize the importance of individual patches by using the result of self-attention. For our task of font identification, attention rollout will give higher (or lower) attention to the patches which are important (or unimportant) for the identification. \par
For local style awareness, it is very important that the patch-wise attention map with attention rollout realizes a higher spatial resolution than other XAI techniques, such as Grad-Cam~\cite{selvaraju2017grad},  which is a popular XAI to visualize the regions that contribute to the decision in Convolutional Neural Networks (CNN). It is well-known that the spatial resolution by Grad-Cam is very low because it depends on the size of the deepest convolution layer. In contrast, ours has $M\times N$ resolutions, and theoretically, using smaller patches makes $M$ and $N$ larger. In practice, however, using too small patches is not good to describe the local shape. The current resolution of the local style awareness in Fig.~\ref{fig:teaser} is a good compromise between resolution and descriptive power and still finer than Grad-Cam. 
% The details of the attention rollout are described in Appendix A.
% -------------------------------------------------------------------
\subsubsection{Quasi-self-supervised learning}
To train the model of Fig.~\ref{fig:teaser}, we need to give 
ground truth (``Same'' or ``Not'') for each character image pair. This ground truth information, fortunately, can be given without any manual annotation effort. As noted in Section~\ref{sec:introduction}, if we can prepare a set of fonts (say, {\tt Helvetica} and {\tt Optima}), they automatically indicate which character images come from {\tt Helvetica} or {\tt Optima}. Such indications are enough to give the ground truth. Since this framework still needs external information (on preparing font sets), it is not fully self-supervised, which does not require any external information. Therefore, we call it {\em quasi-}self-supervised learning. From a practical viewpoint, however, it is equivalent to self-supervised learning because its annotation cost is zero after font set preparation.
\par
% -------------------------------------------------------------------
\subsubsection{Implementation details}
% ViTの説明 （パラメータの詳細も)
% attention rolloutの説明
% acc 0.9469416126042632
The transformer encoder in ViT follows the implementation of ViT~\cite{dosovitskiy2020image} pretrained by ImageNet-21K.
The classifier in Fig.~\ref{fig:teaser} consists of two fully connected layers. The numbers of layers and heads in the transformer are 12.
The class token is a 768-dimensional vector.
The size of an input image and the patch size are $224\times224$ and $16\times16$, respectively. (Therefore, $M=N=14$.)
Batch size and learning rate are set at $64$ and $10^{-5}$, respectively.
We use Adam for the optimizer and cross-entropy loss for the loss function.
% -------------------------------------------------------------------
\subsection{Qualitative evaluations of local style awareness\label{sec:eval-awareness}}
% -------------------------------------------------------------------
\subsubsection{Dataset}
% topic: google fontsを使います (以下，データセット収集の方法)
% google fontsの各フォントの METADATA.pd からフォントファミリー名，フォントカテゴリ ，文字サブセット情報を取得
% 文字サブセットに‘ltain’を含むフォントのみ選定
% フォントファミリーでtrain, valid, testを分割
% フォントファミリーに含まれている italic やboldなどを展開（train,valid,testそれぞれのデータに使用）
% それぞれのttfから実際に0~9,A~Z,a~zの文字をレンダリング可能なもののみ使用
% 実験に使用した文字種類 大文字A~Z
We used the font dataset from Google Fonts \footnote{\url{https://github.com/google/fonts}} as follows.
First, using metadata, we obtained font family name, category name\footnote{We use four categories of fonts included in Google Fonts. In more detail, there are 1,283 Sans-Serif fonts, 630 Serif fonts, 457 Display (i.e., decorative) fonts, and 203 Handwriting fonts.}, and a character subset, which shows languages included in each font.
We chose the fonts with a character subset of ``Latin'' and discarded the others. We also discarded incomplete fonts.
Then, we divided the font into a training, validation, and testing set to 8:1:1. 
During division, we did our best to avoid very similar fonts in different sets by checking font family names. 
As a result, we prepared 2,094 fonts, 230 fonts, and 249 fonts for the training, validation, and testing sets, respectively. For simplicity (by avoiding the disturbances of small caps.), we only used 26 capital letters in the following experiments.\par
The original font data is the vector format (TTF); therefore, we render it to bitmap images of $224\times 224$  pixels with a margin of 5 pixels to use the experiment of font identification.
In font generation, we resize these images to $64\times 64$ or $80\times 80$ to adapt to the experimental setting for each baseline of the font generation models.
% 
% -------------------------------------------------------------------
\subsubsection{Comparative models}
% topic: 比較手法を２つ準備する
% 1．font category classification (ViT) + Attention rollout
% 2. font identification (CNN) + Grad-CAM
Although there is neither similar work nor baseline, we design two comparative models for evaluating local style awareness in font images. 
\begin{itemize}
\item One is a ViT trained for the font category classification task (instead of font identification). Then, we obtain its attention map by attention rollout.  
Font category classification is a task that classifies the input font image into one of the four font categories, ``Serif,'' ``Sans Serif,'' ``Handwriting'' and ``Display.'' These categories are given in Google Fonts. ViT pretrained by ImageNet-21K is fine-tuned for font category classification. 
The hyper-parameters in the model are the same as the ones in the font identification in Section~\ref {sec:tech-details}.
\item The other is CNN (instead of ViT) trained for the font identification task. Then, we obtain its Grad-CAM.
We employed ResNet-18~\cite{he2016deep} as the CNN. The way of making pairs in the training phase is the same as the identification by ViT.
\end{itemize}
The test accuracy of font identification by ViT, font category classification by ViT, and font identification by CNN are 94.69 \%, 86.38\%, and 94.59\%, respectively. Note that font category classification is difficult because of fuzzy class boundaries between four categories. (Especially the boundary between Sans-Serif and Display is often confusing.)
% -------------------------------------------------------------------
\subsubsection{Visualization of local style awareness\label{sec:visualization}}
\begin{figure}[t]
    \centering
    \includegraphics[width=1.\linewidth]{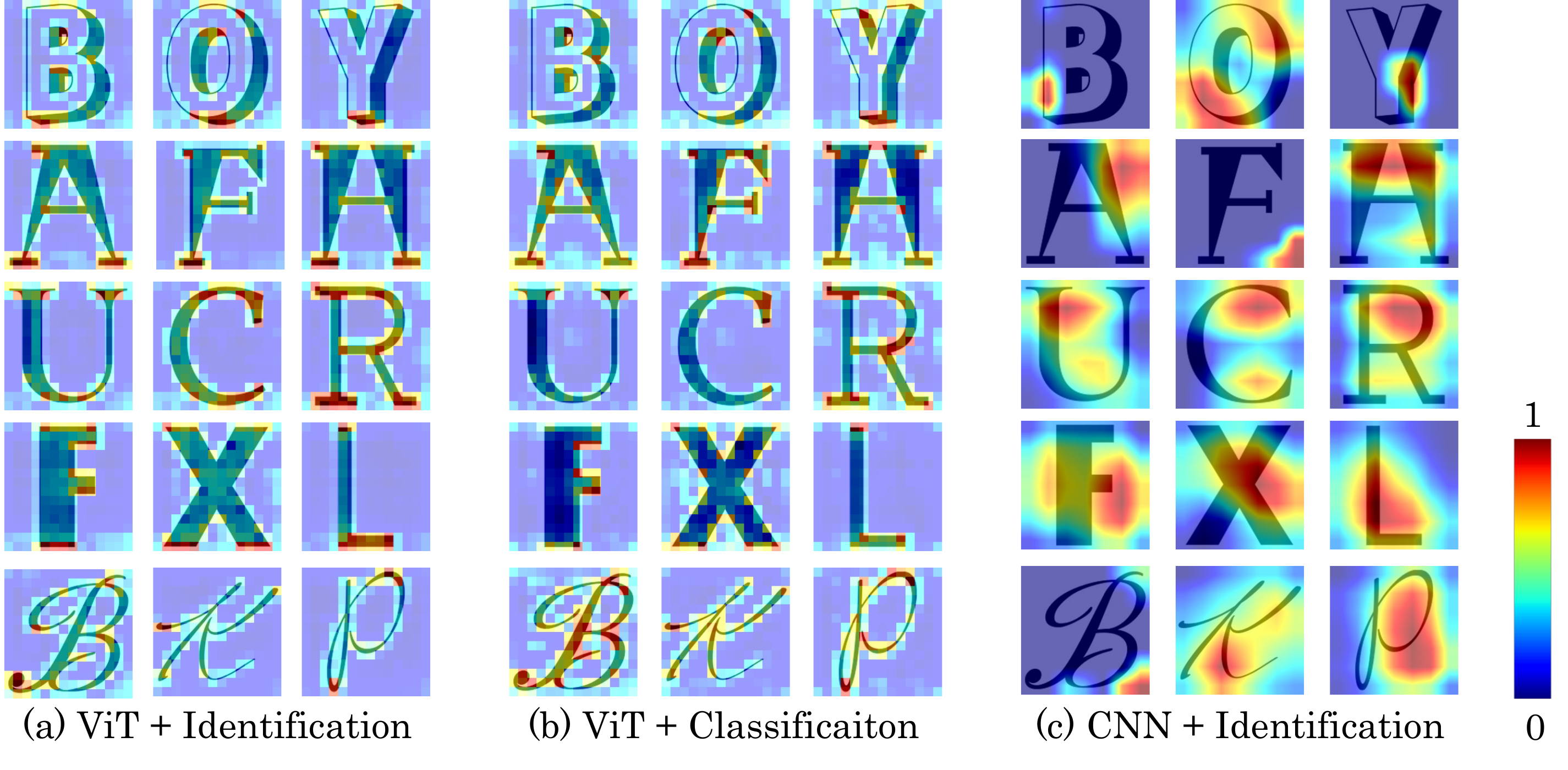}
    \caption{Visualization of local style awareness. Red regions show strong attention, and blue regions show weak attention.}
    \label{fig:local-style-awareness}
\end{figure}
Fig.~\ref{fig:local-style-awareness}~(a) visualizes local style awareness obtained by the proposed model. In the first row, strong attention is found in a part of shadows. (Note that the bottom part of these characters are shadows.) We can also see the consistency of attention to serif parts in the second row.  In the third row, strong attention is found not only in the serif parts but also curves in `U' and `C.'  For the sans-serif fonts in the fourth row, attention is found at the bottom, where the stroke thickness and straightness are clearly represented.  The fonts in handwriting style in the fifth row show attention around their curvy stroke ends and intersection parts.
From observing those maps, our attention maps, showing local stroke awareness, can find local parts that represent font-specific local structures.\par
The comparison between (a) and (b) suggests how the font identification task is more suitable for local style awareness than the font category classification task. In the second row of (b), the comparative model of category classification could capture serifs because the model is trained to discriminate serif fonts from others. However, except for the serif parts, the comparable model of (b) often fails to catch representative local parts. For example, for the samples in the third row, this model totally ignores the curves because the curves are not important for the current category classification task. Similarly, in the fourth row, the model also seems to ignore the thickness and straightness --- it focuses on the corners to check the existence of serifs. To summarize these observations, this model mainly focuses on the corners to discriminate between serifs and sans-serifs and thus is rather insensitive to other local parts representing the unique structure specific to the font. In the next section, we will see how this comparative model captures different style features from ours.\par
The differences between ViT (a)(b) and Grad-CAM (c) in their resolution and accuracy are obvious. As expected, the map by Grad-CAM is very coarse and difficult to understand the important local parts for representing font styles.
Moreover, the map by CNN shows strong attention not only to the character region but also to the background region --- although the background regions are often important for specifying font styles, the Grad-CAM highlight on `F' and `B' seems irrelevant to describe the font style.\par
%

% -------------------------------------------------------------------
\subsubsection{Distributions of local style features}
\begin{figure}[t]
    \centering
    \includegraphics[width=0.9\linewidth]{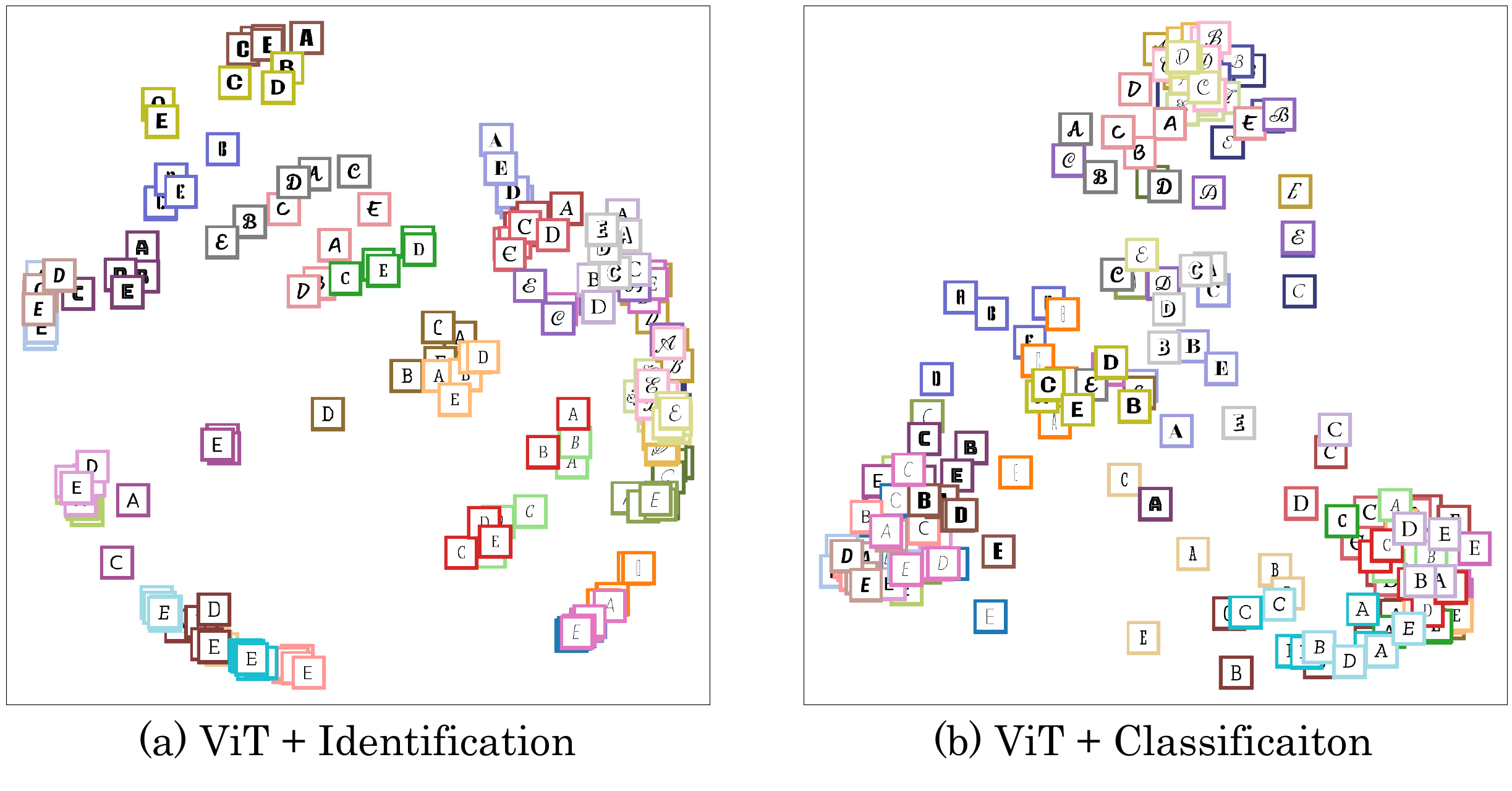}
    \caption{Distributions of class tokens (i.e., style features) by two-dimensional PCA. The same box color indicates the same font.}
    \label{fig:visualization}
\end{figure}

For a further comparison between the proposed model and the comparative model trained for the category classification, 
we visualize the distributions of their class tokens, that is, style features, by ViT. Fig.~\ref{fig:visualization} shows their distributions for the samples of five alphabets from `A' to `E' by two-dimensional PCA. The comparative observation of (a) and (b) shows that the characters from the same font are more clustered in (a) than (b). Consequently, our model based on font identification is more sensitive to the style and can ignore the whole letter shapes. 

% ===================================================================
\section{Boosting Font Generation Quality by Local Style Awareness}
% ===================================================================
% -------------------------------------------------------------------
\subsection{Methodology}
In this section, we utilize local style awareness in font generation tasks for realizing local style-aware font generation.
As shown in Fig.~\ref{fig:gen_example}, local style awareness represents the importance of individual patches. Thus, we can use it for weighting the reconstruction loss function in various font generation models. Adding the weight of local style awareness will contribute to a more accurate reproduction of the important local parts in the generated images.\par
L1 loss weighted by local style awareness $W$ is as follows:
\begin{equation}\label{eq:weighted_L1}
    L =  || (W + \alpha ) \odot (X_{\rm{gt}} - X_{\rm{gen}}) ||_1,
\end{equation}
where $\odot$ is an element-wise product, and $\alpha$ is a constant value for computing normal L1 loss.
We set $\alpha=0.1$ in the following experiment.
Additionally, $X_{\rm{gt}}$ and $X_{\rm{gen}}$ indicate a ground truth image and a generated image, respectively.
% Note that the derivation of local style awareness $W$ by attention rollout is described in Appendix A.
% -------------------------------------------------------------------
\subsection{Quantitative and qualitative evaluation experiments}
\subsubsection{Few-shot font generation}
We evaluate the usefulness of our proposed loss through few-shot font generation. Few-shot font generation is the task that accepts a few source images and content images (or a content label), and then the style of source images is transferred into the content.\par
In the training phase of few-shot font generation, almost all of the models optimize a reconstruction loss between generated image $X_\mathrm{gen}$ and the target image (\textit{e.g.,} ground truth) $X_\mathrm{gt}$.
Therefore, introducing the proposed reconstruction loss into the font generation model is very straightforward.\par
In this experiment, we use the same dataset used in Section~\ref{sec:eval-awareness}.
We set the image size to $64\times64$ or $80\times80$ according to the experimental setting for each baseline of the font generation models. To this end, we resize the attention maps to the same size as each input image. \par
% -------------------------------------------------------------------
\subsubsection{Three baseline models of few-shot font generation}~\label{sec:baseline}
We picked up three baseline models\footnote{There are indeed newer font generation methods, and our model can be introduced even to them.
Since they have rather complex structures, which might obfuscate the effects of different loss functions (original, L1, and ours), we did not use them.} for few-shot font generation and observed the usefulness of the proposed loss for them.

\begin{itemize}
\item {\bf FANnet}~\cite{roy2020stefann} is the font generation model that can edit the characters while retaining the font style of a source image.
In more detail, FANnet accepts an image of the source font and a one-hot vector corresponding to the character label and then generates the characters with the same style as the source font.
In the training phase, it employs L1 loss for reconstruction loss.
When we conduct the few-shot font generation, we use average features extracted from source images.
\item {\bf EMD}~\cite{zhang2018separating,zhang2020unified} is the style transfer model for font style and has often been used as a baseline of the font generation task. EMD accepts content images; therefore, we fix the content images to a simple sans-serif font in the evaluation. EMD employs L1 loss weighted by the character regions to consider the imbalance between background and character regions.
We compare the loss with ours in our experiments.
\item {\bf Srivatsan \textit{et al.}}~\cite{srivatsan2019deep} proposed a font generation model that disentangles content from style in font images and combines them. They optimized the model by projecting an image onto the frequency using the Discrete Cosine Transform (DCT-II) instead of directly reconstructing an image.
Specifically, they impose a Cauchy distribution which is heavy-tailed distribution in the projected space to generate sharper images.
We also compare the loss with our proposed one.
Note that this model includes a loss function for disentangling the style. Therefore, reconstruction loss is not dominant compared with the other methods.
\end{itemize}
% -------------------------------------------------------------------
\subsubsection{Evaluation metrics}%(weighted reconstruction lossを用いた評価も実施)
We evaluate the quality of font generation with various evaluation metrics\footnote{Some metrics might take infinite value when the generated image becomes empty. Therefore, we exclude such images.}.
L1, LPIPS (Learned Perceptual Image Patch Similarity), and SSIM (Structural Similarity) are evaluation metrics commonly used for font generation. Hausdorff distance and IoU are used for the quantitative evaluation of several font generations~\cite{wang2020attribute2font,kang2022shared,wen2021zigan}.
When we calculate the Hausdorff distance and IoU, we binarize the image by Otsu's method.
In the Hausdorff distance, we conduct canny edge detection as preprocessing.
Additionally, we use Pseudo Hamming Distance (PHD)~\cite{uchida2015exploring}, an evaluation metric, to calculate the similarity between fonts. PHD might be the most appropriate way to evaluate font styles among the above metrics because PHD can directly evaluate the difference between two shapes.
Hausdorff distance also directly evaluates the difference.
Roughly speaking, PHD evaluates an average difference over all shape contours, whereas Hausdorff distance evaluates the maximum difference.
Consequently, it is sensitive to slight font shape differences and has been used for evaluating the similarity between font images.
We also evaluate the quality of font generation using our loss function Eq.~\ref{eq:weighted_L1} to set $\alpha = 0$ and call it weighted L1.\par
\begin{table}[t]
    \centering
    \caption{Quantitative evaluation of few-shot font generation. In this experiment, we use five style images. The metrics in \textcolor{magenta}{magenta} are expected to be more sensitive to local differences. The loss of ``original'' is the loss function proposed in each baseline model.}
    \scalebox{0.9}{
    \begin{tabular}{c|c|r|r|r|r|r|r|rr}\hline
         Baseline & loss & L1 $\downarrow$& weighted L1$\downarrow$ & \textcolor{magenta}{Hausdorff}$\downarrow$ & \textcolor{magenta}{PHD}$\downarrow$ & LPIPS$\downarrow$& IoU $\uparrow$ & SSIM$\uparrow$& \\\hline
         FANnet~\cite{roy2020stefann} & L1 & 0.0855 & 0.0347 & 7.172 & 1083 & 0.1542 & 0.6503 & 0.6810 \\
          & ours &\textbf{0.0843} & \textbf{0.0319} & \textbf{5.602} & \textbf{803} & \textbf{0.1294} & \textbf{0.6764}& \textbf{0.6896} \\\hline
         EMD~\cite{zhang2018separating,zhang2020unified} & original&  \textbf{0.0916} & 0.0364 & 8.997 & 2049 & \textbf{0.1523} & 0.6404 & \textbf{0.6829} \\
         & L1 & 0.0938 & 0.0376 & 9.267 & 2139 & 0.1564 & 0.6306 & 0.6794 \\
         & ours & 0.0988 & \textbf{0.0358} & \textbf{8.600} & \textbf{2013} & 0.1543 & \textbf{0.6434} & 0.6666 \\\hline
         Srivatsan \textit{et al.}~\cite{srivatsan2019deep} & original &  0.1219 & 0.0429 & 6.0439 & 1026 & 0.2182 & \textbf{0.6486} & 0.6182 \\
         & L1 &  \textbf{0.0901} & \textbf{0.0362} & 6.866 & 990 & 0.1339 & 0.6335 & \textbf{0.6735}\\
         & ours & 0.1007 & 0.0378 & \textbf{5.699} & \textbf{888} & \textbf{0.1130} &0.6307 & 0.6417\\\hline
    \end{tabular}}
    \label{tab:res}
\end{table}

% -------------------------------------------------------------------
\subsubsection{Quantitative evaluation}
As shown in Table~\ref{tab:res}, our loss could improve all evaluation metrics for FANnet. FANnet is a simple model; therefore, our loss directly improves the font generation quality.
In EMD, ours is better than the others in more than half of the metrics. Especially, ours is best in Hausdorff distance and PHD. These two metrics are more sensitive to the little difference between the images than L1 loss.
This indicates that our loss contributes to generating fonts keeping detailed styles more than the others.
Original loss takes into the imbalance between character regions and background regions. However, for sustaining the font style, using local style awareness for font generation is more effective than the original one.\par
In Srivatsan \textit{et al.}, ours is better than the other model in several metrics. 
In particular, ours is much better than the original loss function in almost all metrics.
Ours is worse than the L1 loss in several evaluation metrics (such as weighted L1 and SSIM).
This model includes not only reconstruction loss but also a loss function for disentangling the style.
The balance between the loss function and reconstruction loss is crucial. Therefore, the order difference between ours and the L1 loss might be one of the reasons for the lower results than the simple L1 loss. It is a limitation of our loss to tune the hyper-parameter (e.g., weight between loss).\par
Through the experiment of all three baseline models, the results by our loss tend to be better in \textcolor{magenta}{Hausdorff} distance and \textcolor{magenta}{PHD}, as shown in Table~\ref{tab:res}.
This indicates that our loss contributes to improving font generation quality in detailed font styles because these losses are sensitive to differences between images, and especially, PHD is a metric to evaluate the similarity between fonts.
Note that our loss aims to sustain the detailed local styles in font generation; therefore, seeing a clear improvement in font generation by using our loss might be difficult in the other evaluation metrics.\par
\begin{figure}[t]
    \centering
    \includegraphics[width=0.9\linewidth]{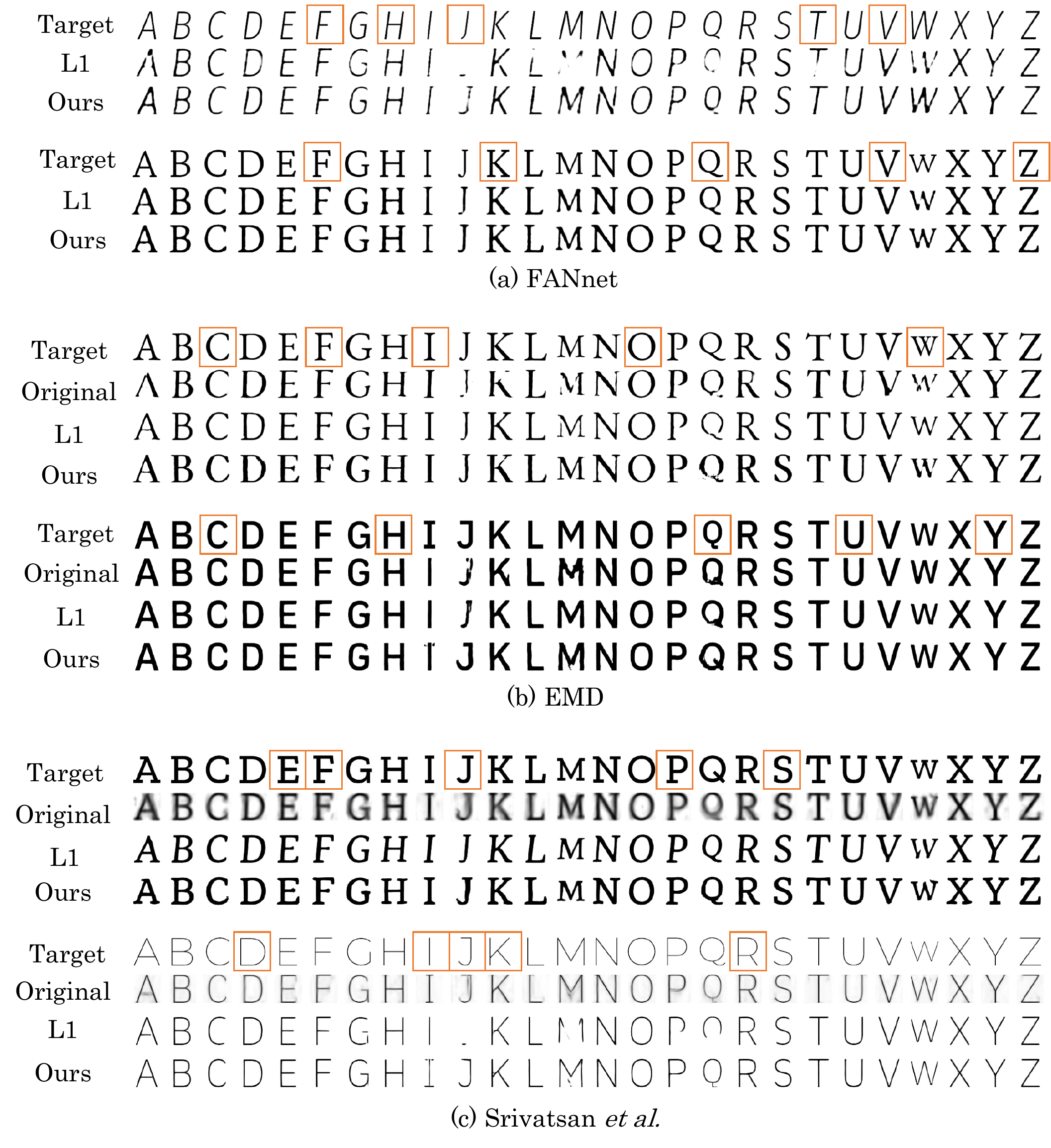}\\[-3mm]
    \caption{Results of few-shot font generation by each baseline model and loss. Orange boxes show the source, which is used for extracting font styles. ``Target'' shows the ground truth.}
    \label{fig:gen}
\end{figure}

However, in some metrics, degradations are caused by two reasons. The first reason is the characteristics of evaluation metrics. For example, we sometimes have better (low) L1 scores when the font generation model generates empty images than generating deformed font images. The second reason is the limitation of our loss function. Our loss function focuses on local shapes representing the style; this implies that some parts unimportant for the style sometimes become noisy. A typical case is 'J' in the first example of Fig.\ref{fig:gen} (c), whose stroke width is not constant. \par

% -------------------------------------------------------------------
\subsubsection{Qualitative evaluation}\label{sec:qualitative}
Fig.~\ref{fig:gen} shows the font generation examples by each baseline model and loss. To generate fonts, we use five source images marked by orange boxes. The source images are chosen randomly.\par
In the first example in FANnet~(a), the L1 loss tends to defect to thin strokes.
Especially, `A,' `J,' and `M' are likely to defect their strokes.
However, ours is effective in fonts with thin strokes.
This is because our loss is correctly weighting to the local style awareness of the font with thin strokes. 
The second example in (a) shows that ours has serif parts more clearly than L1, especially `E,' `F,' and `T.'
From this example, our loss effectively generates font with keeping the local style.\par
In the first example of EMD~(b), the original can not generate serif parts precisely, and some images defect the strokes.
L1 can not capture the stress of stroke width (e.g., `C' and `G'). In contrast, ours can clearly generate fonts with keeping its serif style. In the second example, there is not much difference between the generated fonts. However, we emphasize that only ours can generate `J' correctly.
This result comes from our loss functions advantage that can pay attention to local style awareness.\par
In Srivatsan \textit{et al.}~(c), the first examples show that ours can generate fonts sustaining the detailed serif parts, especially the top serif of `A.' This trend can be seen in ours.
Local style awareness contributes to generating the serif parts like the above `J.' The second example shows that ours can generate thin fonts than L1.  This trend is the same as in (a).
The original method also generates the images; however, several images have a blurry noise.  
% ===================================================================
\section{Conclusion}
% ===================================================================
This paper proposed local style awareness, which represents important shapes for particular font styles. Local style awareness is acquired by solving a font identification task in a contrastive learning scheme. This task is solved very efficiently in a quasi-self-supervised learning manner where no manual annotation is necessary. In other words, we can get local style awareness without human efforts. Our model is based on ViT instead of CNN because ViT and its attention mechanism help us to have finer local style awareness that can catch a small style structure such as serifs.\par
As an application task, we utilized local style awareness in few-shot 
font generation to generate font images whose local structures are 
realized more accurately. In this application, we simply use the local style awareness as the weight for the reconstruction loss function; this simplicity allows us to use the local style awareness in various state-of-the-art font generation models.
In our experiments, we prove quantitatively and qualitatively that our loss could improve the performance of three baseline models. \par
In future work, we will obtain local style awareness of other languages and apply them to a font generation task in the language. Additionally, we will apply the awareness to other tasks, such as style analysis, style transfer, and style domain adaptation.\par

\section*{Acknowledgment}
This work was supported in part by JST, the establishment of university fellowships towards the creation of science technology innovation, Grant Number JPMJFS2132, JSPS KAKENHI Grant Number JP22H00540, and JST ACT-X Grant Number JPMJAX22AD.
% This work was supported by JST, ACT-X Grant Number JPMJAX22AD, and the establishment of university fellowships towards the creation of science technology innovation, Grant Number JPMJFS2132, Japan.

\bibliographystyle{splncs04}
\bibliography{ref}
\end{document}